\documentclass[sigconf]{acmart}

\usepackage{booktabs} 

\usepackage{epsfig}
\usepackage{algorithm}  
\usepackage{algpseudocode}
\algtext*{EndWhile}
\algtext*{EndIf}
\algtext*{EndFor}
\algtext*{EndProcedure}
\algtext*{EndFunction}
\usepackage{soul,color}
\usepackage{enumerate}
\usepackage{multirow}
\usepackage[dash,dot]{dashundergaps}
\usepackage{amsmath}
\newcommand{\Tanh}{\mathrm{Tanh}}
\newcommand{\eat}[1]{}
\usepackage[normalem]{ulem}
\setcopyright{none}
\renewcommand\footnotetextcopyrightpermission[1]{}
\begin{document}
\title{SpatialNLI: A Spatial Domain Natural Language Interface to Databases Using Spatial Comprehension}

\author{Jingjing Li}\authornote{Both authors contributed equally to this research.}
\affiliation{%
   \institution{
  Auburn University}
}
\email{jingjingli@auburn.edu}

\author{Wenlu Wang}\authornotemark[1]
\affiliation{%
   \institution{
  Auburn University}
}
\email{wenluwang@auburn.edu}

\author{Wei-Shinn Ku}
\affiliation{%
   \institution{
  Auburn University}
}
\email{weishinn@auburn.edu}

\author{Yingtao Tian}
\affiliation{%
   \institution{
  Stony Brook University}
}
\email{yittian@cs.stonybrook.edu}

\author{Haixun Wang}
\affiliation{%
   \institution{
  WeWork Research}
}
\email{haixun.wang@wework.com}

\begin{abstract}
A natural language interface (NLI) to databases is an interface that translates a natural language question to a structured query that is executable by database management systems (DBMS). However, an NLI that is trained in the general domain is hard to apply in the spatial domain due to the idiosyncrasy and expressiveness of the spatial questions. Inspired by the machine comprehension model, we propose a spatial comprehension model that is able to recognize the meaning of spatial entities based on the semantics of the context. The spatial semantics learned from the spatial comprehension model is then injected to the natural language question to ease the burden of capturing the spatial-specific semantics. With our spatial comprehension model and information injection, our NLI for the spatial domain, named {\it SpatialNLI}, is able to capture the semantic structure of the question and translate it to the corresponding syntax of an executable query accurately.
We also experimentally ascertain that SpatialNLI outperforms state-of-the-art methods.

\end{abstract}

%
%



\maketitle
\section{Introduction}
Many business applications rely on relational databases. To facilitate the usage of database management systems to the public, 
NLI to databases has been extensively studied \cite{androutsopoulos1995natural,DBLP:conf/ijcnlp/BradIHR17,DBLP:journals/corr/abs-1804-00401,DBLP:conf/sigmod/LiJ14,DBLP:conf/sigmod/LiYJ05,DBLP:conf/iui/PopescuEK03,DBLP:journals/pvldb/SahaFSMMO16,DBLP:conf/acl/IyerKCKZ17,wang2018transfer,wang2019cross}.
Spatial Domain NLI to databases has drawn great attention due to the popularity of spatial applications~\cite{wang2017efficient, wang2019scalable,wang2017dynamic,wang2017recommendation}. 
An intuitive solution is to adopt existing NLI in general databases to the spatial domain.
However, due to the idiosyncrasy and expressiveness of the spatial semantics, it is unfeasible to adopt general NLI for the spatial domain directly. 
The challenge of adopting the existing general domain NLI to spatial domain lies to harnessing the expressiveness of spatial semantics.
The expressiveness of spatial semantics can be justified based on the following observations~\cite{spatial-semantics}:

\begin{figure}[h!]
    \centering
     \begin{tabular}{l c}
      \multicolumn{2}{c}{The meaning of spatial phrase ``{\it Mississippi}''}\\
    \hline
    {\it How many rivers does \textbf{Mississippi} have} ?   &state\\
    {\it How many cities does \textbf{Mississippi} run through} ? &river\\
    \\
    \multicolumn{2}{c}{The meaning of spatial phrase ``{\it over}''}\\
    \hline
    {\it How many people walked \textbf{over} the bridge} ? & on\\
    {\it How many birds flew \textbf{over} the bridge} ?  & above\\
   \\
    \multicolumn{2}{c}{The meaning of spatial phrase ``{\it at the back of}''}\\
    \hline
    {\it How many trees are \textbf{at the back of} the building} ? &exterior\\
    {\it How many rooms are \textbf{at the back of} the building} ? &interior\\
    \end{tabular}
    \caption{Three examples show that the spatial semantics is encyclopedic.}
    \label{fig:spaial_examples}
\end{figure}

The examples as mentioned earlier show that the same spatial phrase in different questions embodies divergent senses expressing divergent query intentions.
In the first two questions, ``{\it Mississippi}''
as a name can refer to either a state or a river, depending on the context where it is mentioned. In this example, the type of word  ``{\it Mississippi}'' depends on the verb located after the name (``{\it have}'' or ``{\it run through}''). 
In the second two questions, the preposition ``{\it over}'' means either a superior position or on the surface. Its spatial meaning depends on the verb located before the preposition (``{\it walk}'' or ``{\it fly}'').
In the last two questions, the prepositional phrase ``{\it at the back of}'' means either outside the building or inside the building,
which depends on the noun before the prepositional phrase (``{\it tree}'' or ``{\it room}''). Such contextually dependent spatial semantics raises serious challenges for NLI to spatial domain databases. For instance, in the third example, if there are two spatial tables (one for the interior architecture of a building and one for the surroundings of a building),
a wrongly comprehended spatial semantics would cause the NLI to query a wrong table.
In general, spatial semantic understanding relies heavily on its contextual interpretation. 

Existing works of NLI rely on conventional grammar-based methods or neural network-based methods. The former line of existing work uses predefined templates or manually designed features, which has the lower-transfer ability, thus confined in its specific dataset.
The latter line of existing work uses grammar embedded neural networks. Embedding grammar into a model relies on converting the process of generating a sequence of tokens to the task of generating a sequence of actions that expands a syntax tree. Converting word space
to action space will inevitably introduce transformation error, which can not guarantee overall accuracy. To the best of our knowledge, the state-of-the-art Syntax-based method TRANX~\cite{DBLP:conf/emnlp/YinN18} achieves an accuracy of $88.2\%$ (Geoquery dataset) which is lower than our accuracy.

The aforementioned observations and survey inspired us to propose a Spatial Domain NLI that is able to support the idiosyncrasy of spatial semantics.
Inspired by the NLI in~\cite{wang2018transfer}, we 
propose a strategy to address the ambiguity of spatial meaning (mentioned in Figure~\ref{fig:spaial_examples}) and data sparsity problem by feeding necessary spatial semantics to the deep model. Here ambiguous spatial phrases are those that can not be uniquely identified by the schema.
By feeding external spatial semantics, our NLI is able to support various spatial questions even when it has not seen similar semantics in the training set. 
The extra spatial semantics is recognized by our external spatial comprehension model, whose functionality is to recognize pre-defined spatial semantics.

We propose to capture spatial semantics using an external spatial comprehension model, where the interpretation of each word is based on the attentive combination of the context.
We then complete our NLI model using a sequence-to-sequence (seq2seq) translation, which is not only able to achieve grammar correctness but also robust with data sparsity problem.
Our fundamental strategy is to separate the tasks of NLI to (1) learning semantic structure of a natural language question, and (2) learning the spatial semantics of a spatial question.

The necessity of the external spatial comprehension model is due to the seq2seq translation model's failure to capture all the spatial semantics while learning the structure of the question.
In our design, Task (1) is assigned to the seq2seq model, while an external spatial comprehension model is in charge of Task (2).
We propose our spatial comprehension model as a bi-directional attentive workflow~\cite{DBLP:conf/iclr/SeoKFH17, DBLP:conf/iclr/Wang017a}, where the attentive spatial phrases of the input are enclosed with special symbols.

Our strategy is a general-purpose automatic solution that only relies on database content, an external model, a seq2seq model, and a minimum amount of human knowledge.
To the best of our knowledge, we are the first to use an external spatial semantic understanding model to enhance the performance of the main seq2seq model. Our solution not only addresses the problem of data sparsity but also introduces minimum error since the spatial comprehension model achieves an accuracy of $98\%$ and $100\%$ for Geoquery and Restaurant datasets.

Our contribution is described as follows
\begin{itemize}
\item We propose a spatial comprehension model that is able to recognize the meaning (e.g., POI type) of an ambiguous spatial phrase (e.g., POI name) based on contextual interpretation.
\end{itemize}

\begin{itemize}
\item After injecting spatial semantics learned from spatial comprehension into the question, our model outperforms the state-of-the-art.
\item We evaluate our strategies systematically and show that our spatial comprehension model and injection format perform well as expected.
\end{itemize}

\begin{figure*}
    \centering
    \resizebox{0.8\textwidth}{!}{
    \includegraphics{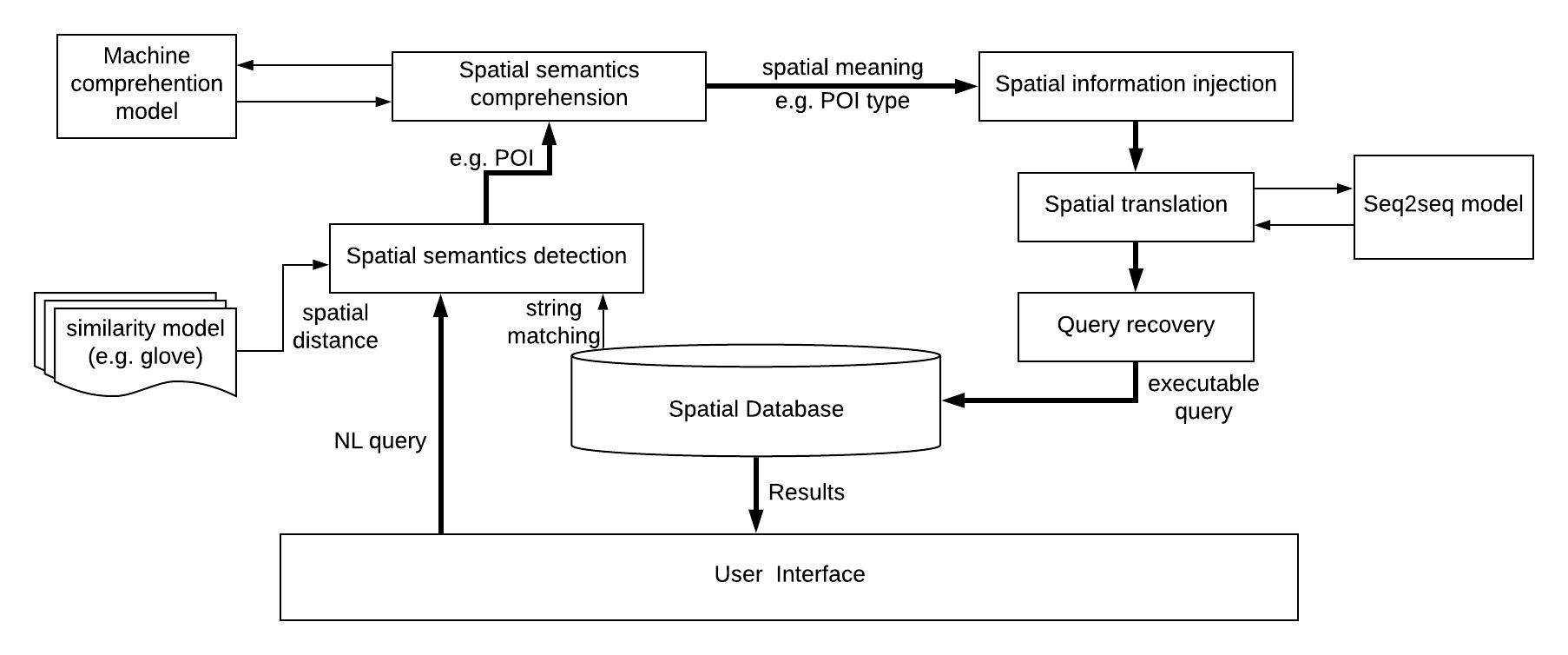}
    }
    \vspace{-15pt}
    \caption{SpatialNLI overview.}
    \label{fig:overview}
    \vspace{-15pt}
\end{figure*}

\section{Related Work}


In recent years, a line of works has been focusing on semantic parsing, which aims at converting natural language utterances to formal meaning representations. ZC05~\cite{DBLP:conf/uai/ZettlemoyerC05}, ZC07~\cite{zettlemoyer2007online},  UBL~\cite{DBLP:conf/emnlp/KwiatkowksiZGS10} and FUBL~\cite{DBLP:conf/emnlp/KwiatkowskiZGS11} induce the specific grammars to make the translation, which defines the meanings of individual words and phrases. KCAZ13 \cite{kwiatkowski2013scaling} and~\cite{DBLP:journals/pvldb/SahaFSMMO16} use ontologies to help form the grammar. \cite{wang2014morpho} uses domain-independent facts to make the translation and ZH15~\cite{DBLP:conf/naacl/ZhaoH15} builds the grammar based on the specific entity type of words. DCS+L~\cite{DBLP:journals/coling/LiangJK13} and~\cite{DBLP:conf/sigmod/LiJ14} introduce tree structure for input natural language to solve this problem. However, most of the conventional methods rely on predefined templates or manually designed features to complete the translation, which is not comparable to ours, as we avoid using such lexicon mapping and predefined templates of prior knowledge in our system. 

Also, some of the work focuses on the Natural Language Interface to Databases (NLIDB) for users to interact with the database without acknowledging the grammar of structured queries executed by the database engine. \cite{androutsopoulos1995natural} first explores this issue with a specific database and concrete examples. \cite{zelle1996learning, DBLP:conf/iui/PopescuEK03, tang2000automated,tang2001using,ge2005statistical} also work on this issue depending on grammars and processes the semantic grammars manually for each individual database. \cite{DBLP:conf/ijcnlp/BradIHR17} and~\cite{DBLP:journals/corr/abs-1804-00401} work on the NLIDB systems requiring large sets of natural language and SQL pairs. \cite{DBLP:conf/sigmod/LiYJ05} and~\cite{DBLP:conf/acl/IyerKCKZ17} present an interface with the help of the feedback from users and PEK03~\cite{DBLP:conf/iui/PopescuEK03} also defines the coverage of the NLIDB system, which is certainly not suitable for all databases. The problem is that the NLIDB study mentioned above is all designed for the general domain and is hard to apply to spatial natural language queries directly without loss of accuracy.


Now deep neural network models have been applied successfully to semantic parsing to exploit the sequential structure on both input and output side. One of them is the Encoder-Decoder model~\cite{kalchbrenner2013recurrent, sutskever2014sequence, cho2014learning}. FKZ18~\cite{DBLP:conf/acl/RadevKZZFRS18} works on translating the input to SQL queries based on the Encoder-Decoder model. TRANX~\cite{DBLP:conf/emnlp/YinN18} and ASN~\cite{ DBLP:conf/acl/RabinovichSK17} construct Decoder-Encoder models with the tree structure. SQL2TREE~\cite{DBLP:conf/acl/DongL16} proposes a seq2seq model based on the Encoder-Decoder architecture and JL16~\cite{DBLP:conf/acl/JiaL16} enhances the performance of seq2seq by adding attention-based copying in the output and implementing data augmentation.  \cite{DBLP:conf/rep4nlp/FanMMD17,wang2018transfer} work on an Encoder-Decoder based transfer learning for semantic parsing. \cite{susanto2017neural} trains one model that is able to parse natural language sentences from multiple different languages and~\cite{DBLP:conf/acl/HerzigB17} exploits the Encoder-Decoder model in different domains. \cite{DBLP:journals/corr/abs-1709-00103} introduces a framework with reinforcement learning to generate SQL queries. Here, we introduce on seq2seq model in our system. Compared with SQL2TREE~\cite{DBLP:conf/acl/DongL16} and JL16~\cite{DBLP:conf/acl/JiaL16}, our model solves the spatial ambiguity problem for the input natural language more efficiently. 

Natural language process for the spatial domain has been observed in literature. \cite{shen2009annotation} annotates the spatial relation in natural language based on the specific annotation schema. \cite{bateman2010linguistic} focuses on spatial ontologies to process the input spatial natural language queries. \cite{kordjamshidi2011relational} maps the objects and spatial relations to formal linguistic terms, which disambiguate the spatial meanings of objects. \cite{khan2013extracting} uses a form of symbolic expressions to extract spatial terms from natural language descriptions to represent spatial features and relations between them. All of them limit the query in the fixed form and have difficulty dealing with different kinds of spatial complex queries. For~\cite{ramalho2018encoding},  it introduces a system that is capable of capturing the semantics of spatial relations in natural language using the neural network. But none of the above gives users an interface to interact with the database.


\section{Challenges of spatial NLI}\label{sec:challenges}
\begin{enumerate}[I.]
    \item Sparse training data. Even though data augmentation is a feasible solution to sparse training data, it is likely that the deep model will be forced to handle unseen questions that are not covered by data augmentation, and required to support transfer learning.
    \item  Spatial semantics ambiguity. A unique feature of spatial questions is its expressiveness in the spatial domain, and a spatial phrase often has an ambiguous meaning. For example, ``{\it Mississippi}'' could be either a state or a river, ``{\it New York}'' could be either a city or a state (taking Figure~\ref{fig:example1} as examples).
\end{enumerate}

\begin{figure}[h!]
    \vspace{-10pt}
    \centering
    \begin{tabular}{c|l}
         & \multicolumn{1}{c}{Question} \\
    \hline
     \textcolor{blue}{State}  &  {\it How many people live in   \textcolor{blue}{Mississippi}?}\\
     \textcolor{blue}{River}  &  {\it How many states does the \textcolor{blue}{Mississippi} run through?}\\
    \hline
      \textcolor{blue}{City}  &  {\it Is \textcolor{blue}{New York} or London bigger?}\\
     \textcolor{blue}{State}  &  {\it What is the capital of  \textcolor{blue}{New York} ?}
    \end{tabular}
    \vspace{-10pt}
    \caption{Spatial POI ambiguity}
    \label{fig:example1}
    \vspace{-5pt}
\end{figure}

Theoretically, a powerful data augmentation should be able to address the first challenge; however, such data augmentation strategy is rare in practice.
Moreover, a seq2seq model is designed to translate a sentence, it is reasonable that it fails to capture the context precisely and infer the correct spatial semantics.
For example, in the question  ``{\it How many rivers does Mississippi have} ?'' (shown in Figure~\ref{fig:poi_injection}), a seq2seq model should be able to understand the context and infer ``{\it Mississippi}'' as a state.
However, since the word ``{\it rivers}'' appears in the question and precedes the word ``{\it Mississippi}'', which means it has a major impact on the prediction when attentive on ``{\it Mississippi}'', in that case, it is highly possible that  \texttt{stateid} will be inferred instead of  \texttt{riverid}.

\begin{figure}[t]
    \centering
    \resizebox{.5\textwidth}{!}{
    \begin{tabular}{|c|c|}
    \toprule
     Question &  {\it How many rivers does Mississippi have} ?\\
    \hline
     Ground &  \multirow{2}{*}{\texttt{answer(A,count(B,(river(B),const(C,\textcolor{blue}{stateid}(Mississippi)),loc(B,C)),A))}}\\
     Truth & \\
    \hline
     Infer & \texttt{answer(A,count(B,(river(B),const(C,\textcolor{blue}{riverid}(Mississippi)),loc(B,C)),A))}\\
    \bottomrule
    \end{tabular}
    }
    \vspace{-10pt}
    \caption{POI type recognition without spatial comprehension model}
    \label{fig:poi_injection}
\end{figure}

Therefore, we propose another deep model for the purpose of spatial semantic understanding; despite the fact that we use the same parse training data, an external model targeted on understanding the context is able to infer the correct spatial semantics (Challenge II) precisely. With the spatial semantics retrieved from the spatial comprehension model, we adopt symbol insertion strategy~\cite{wang2018transfer} to inject external information and help the seq2seq to infer an unseen sample correctly (Challenge I).

\noindent\textbf{Overview} To address the aforementioned challenges, we present our \textit{SpatialNLI} overview shown in Figure~\ref{fig:overview}.
The workflow of our SpatialNLI involves the following steps:
\begin{enumerate}[1.]
    \item Identify ambiguous spatial semantics in the NL query.
    \item Build a spatial comprehension model that is able to understand a spatial-related question semantically.
    \item Injecting spatial semantics retrieved from the spatial comprehension model into the question ($q \longrightarrow q'$).
    \item ``Translating'' the question into a structured query (Lambda expression in our example) ($q' \longrightarrow l'$).
    \item Replace the symbols injected to their original text ($l' \longrightarrow l$).
\end{enumerate}

\section{\textsc{SpatialNLI}}

Since most of the keywords or data elements in spatial queries (e.g., lambda expression) are spatial-related, we propose a strategy to inject latent spatial semantics into the natural language question to help the seq2seq model to capture the semantic meaning of the question.
For example, for the question ``{\it How many rivers does Mississippi have} ?'', its correspondence lambda expression is ``\texttt{\small count(B,(river(B), const(C,stateid(Mississippi)), loc(B,C))}'', which has five keywords ``{\it count}'', ``{\it river}'', ``{\it const}'', ``{\it stateid}'', ``{\it loc}'', and three of them are spatial-related. We will illustrate the workflow of our SpatialNLI with this running example.

Our SpatialNLI model is composed of the following steps (corresponding to Algorithm~\ref{alg:spatialNLI})
\begin{enumerate}[1.]
    \item \textbf{Spatial Semantics Detection}. Having access to GeoSpatial databases, we detect potential keywords or data elements using 1) string match, 2) edit distance, and 3) cosine distance in semantic embedding space (e.g., Glove). In the aforementioned question, ``{\it Mississippi}'' can be detected by comparing against the data in the databases using string match, ``{\it river}'' can be detected by edit distance since ``{\it rivers}'' is in the table. We will detail semantic distance measurement later in Section~\ref{sec:detection}.
    \item \textbf{Spatial Comprehension Model}. For ambiguous spatial phrases, we propose a spatial comprehension model to resolve the ambiguity. As we mentioned in Section~\ref{sec:challenges}, ``{\it Mississippi}'' is an ambiguous POI, and it will be identified as a river type using our spatial comprehension model.
    \item \textbf{Spatial Semantics Injection}. With identified key word ``{\it river}'' and data element ``{\it Mississippi}'' (river name), we inject such information into the question by inserting pre-defined symbols ``{\it How many $\langle k0\rangle$ rivers $\langle eok\rangle$ does \underline{$\langle k1\rangle$ stateid $\langle eok\rangle$} $\langle v0\rangle$ Mississippi $\langle eov\rangle$ have} ?'' 
    \item \textbf{Seq2seq Translation}. We then feed
    the modified question to a seq2seq translation model. In the previous example, the predicted output sequence is ``\texttt{answer(A, count(B, ($\langle k0\rangle$(B), const(C,$\langle k1\rangle$($\langle v0\rangle$)), loc(B,C)), A))}''.
    \item \textbf{Query Recovery}. The generated sequence of the seq2seq model is then recovered to an executable query. Following the previous example, we have ``\texttt{answer(A, count(B,(river(B), const(C, stateid(Mississippi)), loc(B,C)), A))}''.
\end{enumerate}

\begin{algorithm}[t]
\caption{SpatialNLI} 
\begin{algorithmic}[1]
\Function{\textsc{SpatialNLI}}{$q$, $D$, $E$}
    \State $P$, $V$ = \textsc{SpatialMapper}($D$, $q$, $E$);
    \State $q'$, $s2p$ =\textsc{SpatialInjection}($q$, $V$, $P$);
    \State $l'$ = \textsc{Seq2seq}($q'$);
    \State $l$ = \textsc{Recover}($l'$, $s2p$);
    \State \textbf{Return} $l$;
\EndFunction
\end{algorithmic}
\label{alg:spatialNLI}
\end{algorithm}

\subsection{Spatial Semantics Detection}\label{sec:detection}

Even though our major contribution is spatial comprehension, we formally define our strategy to detect keywords and data elements mentioned in the question (denotated as $SpatialMapper$) to keep our work self-contained. 

\vspace{-10pt}
$$P, V = \textsc{SpatialMapper}(D, q, E)$$

The inputs are the GeoSpatial database $D$, a natural language question $q$, and an embedding function $E$ (e.g., Glove). $E$ will change a word to a high-dimensional vector, which represents its location in the embedding space. We collect the table names, column names, and column values from $D$, thus $D$ refers to a collection of entities in our spatial mapper. The table names (e.g., river) and column names (e.g., river length) in $D$ are potential keywords of executable queries (e.g., Lambda expression), and column values (e.g., Mississippi) are potentially data elements that might be mentioned in executable queries. The detail of the algorithm is presented in Algorithm~\ref{alg:spatial_mapper}, in the aforementioned example, $P=[\langle rivers, river\rangle, \langle Mississippi, Mississippi\rangle]$, since ``{\it Mississippi}'' is detected by exact string match, and ``{\it rivers}'' (in $q$) and ``{\it river}'' (in $D$) has a small edit distance. 
We define semantic distance measurement as
$$\mathrm{semantic\_distance}(a, b) = 1- \frac{E(a)\cdot E(b)}{||E(a)||_2 ||E(b)||_2}$$
The semantic distance is also the spatial distance in the embedding space.
If any operand is a phrase which comprises multiple tokens, for example, $A$ is a list of tokens, we define
$E(A) = \mathrm{avg}_{a\in A}\big(E(a)\big)$.

Taking question ``{\it Where is the lowest spot in Iowa}?'' as an example, its corresponding logic form is \texttt{answer(A,lowest(A,(place(A),} \texttt{loc(A,B),const(B,stateid(Iowa))))}, ``{\it spot}'' in the NL question is matched to keyword ``{\it place}'' since $\mathrm{semantic\_distance}(place,spot)$ $< 0.368$, which is relatively small.

\begin{algorithm}[t]
\caption{Spatial Semantics Mapper} 
\begin{algorithmic}[1]
\Function{\textsc{SpatialMapper}}{$D,q,E$}
     \State $P$ = $\emptyset$; \Comment{Spatial semantics matching pairs.}
     \State $V$ = $\emptyset$; \Comment{Spatial values with its semantic meaning.}
     \For{$k$ in $K..1$}\Comment{Iterating from $K$-gram to $1$-gram}
     \ForAll {$p_q$ $\in$ $k$-gram of $q$} 
        \ForAll{$c \in D$}
            \If {$p_q == c$  or $\mathrm{semantic\_distance}(p_q, c) < \tau_{sem}$ or edit\_distance($p_q, c$) < $\tau_{ed}$} \Comment{$\tau_{ed}$ is the threshold for edit distance. $\tau_{sem}$ is the threshold for semantic distance.}
                 \State $P$.add($\langle p_q, c\rangle$);
                  \If {$c$ is a column value}\label{line:amb0}
                    \ForAll{table $tb$ that has $c$}
                        \State $p_q.types$.add($tb$)
                    \EndFor
                    \State $V$.add($p_q$)\label{line:amb1}
                  \EndIf
            \EndIf
        \EndFor
     \EndFor
     \EndFor
     \State \textbf{Return} $P$, $V$;
\EndFunction
\end{algorithmic}
\label{alg:spatial_mapper}
\end{algorithm}

We care about the spatial phrases that have semantic ambiguities (e.g., {\it Mississippi}). An intuitive solution is to use pre-collected human knowledge. However, to devise an automatic and intelligent approach, we propose using Geospatial database; for example, we discover that ``{\it Mississippi}'' is an ambiguous value by simply searching for this phrase in the database, and it appears in two tables~\textsc{River} and~\textsc{State}. In Algorithm~\ref{alg:spatial_mapper} Line~\ref{line:amb0}-\ref{line:amb1}, if a phrase appears in multiple tables, we collect all the ambiguous information in $V$. For example, in $V$, ``{\it Mississippi}''.$types$ = [\textsc{River}, \textsc{State}], ``{\it New York}''.$types$ = [\textsc{City}, \textsc{State}] and ``{\it Alabama}''.$types$ = [\textsc{State}]. In other words, for a spatial phrase that is a value, we collect the tables it belongs to and stored in $V$. For most of the Geospatial databases, the table names are able to represent the meaning of the value. In the question ``{\it How many rivers does Mississippi have}?'', $V$ = [``{\it Mississippi}''] and ``{\it Mississippi}''.$types$ = [\textsc{River}, \textsc{State}].

It is worth noticing, we use minimum human knowledge to cover phrase mapping that is not covered by Glove. For example, in a question where ``{\it population per $km^2$}'' refers to ``{\it population density}'',
such mapping is not easy to be covered by Glove or a deep model, thus human knowledge is necessary. However, such cases are rare in practice, and we only require minimum human knowledge.

\subsection{Spatial Comprehension}

\begin{figure*}
    \centering
    \resizebox{0.7\textwidth}{!}{
    \includegraphics{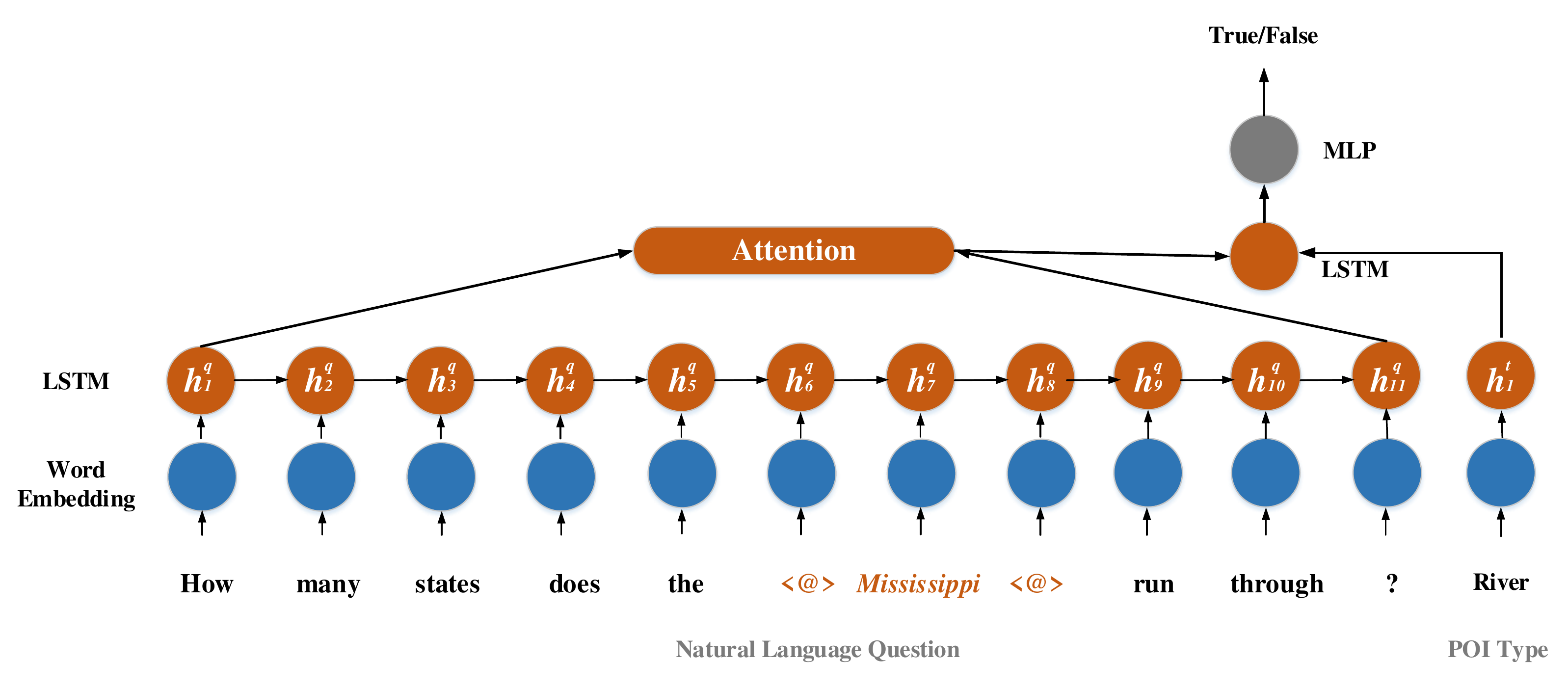}
    }
    \vspace{-15pt}
    \caption{Spatial Comprehension Model}
    \label{fig:spatial-comprehension-model}
\end{figure*}

A critical challenge in understanding the spatial question is the meaning of an ambiguous phrase, such as a point of interest (POI).
For example in Figure~\ref{fig:example1}, where ``{\it Mississippi}' could be a river or a state,
we have to differentiate its meaning by the context and understanding the semantic meaning of the question.
In the question ``How many people live in {\it Mississippi} ?'', we would interpret ``{\it Mississippi}'' as a state name,
and in the question ``How many states does the {\it Mississippi} run through ?'', people would understand ``{\it Mississippi}'' is referring to a river.

With large training corps, a deep model might be able to capture that information; however,
existing spatial question answer data sets are inadequate and sparse due to the difficulty in collecting ground truth. To address this challenge,
we propose a principle method to enable semantic understanding of spatial questions using sparse training data, which relieves the burden of collecting large training sets.
By our definition, \textit{Spatial Comprehension} is spatial semantic understanding using machine comprehension.
Our strategy is to exploit pre-trained Glove embedding to understand spatial keywords in the question first, then use a seq2seq machine comprehension model to learn the semantic meaning of the question (context) without the burden of extracting the spatial relations.

\textbf{Model Structure}
Our spatial comprehension model is designed to understand the ``meaning'' (e.g., type of POI) of an ambiguous spatial phrase mentioned in the question based on its context.

Inspired by the machine comprehension model using an attention flow~\cite{DBLP:conf/iclr/SeoKFH17}, we propose our spatial comprehension model composed of two stacked LSTM layers on each input with another shared attentive LSTM layer. 
The design of the bi-directional attentive workflow~\cite{DBLP:conf/iclr/SeoKFH17} is to answer a question given a premise -- i.e., locate the sentences in the premise that is most relevant to the answer of the question. The task of the machine comprehension is different from ours; however, they do share the same strategy: understanding one of them (question and premise) semantically based on the context of the other. So we use LSTM to pre-process our NL question and the possible meaning of the ambiguous spatial phrase separately, and conduct an attentive workflow over the hidden states of the question (shown in Figure~\ref{fig:spatial-comprehension-model}).
Also inspired by~\cite{wang2018transfer}, we enclose the ambiguous spatial phrase in special symbols to indicate it has more influence than the other tokens in the question.

We denote a question as $q$ = $[q_1,...,q_n]$ and the meaning of the ambiguous spatial phrase (mentioned in the question), such as a POI type, as $t$ = $[t_1, ..., t_m]$, both of which are fed to a word embedding layer $\phi$ (initialized with GloVe~\cite{glove}). On top of that, we use a LSTM layer to capture the hidden states of each time step $i$.


We build the same structure for both $q$ and $t$.
We denote the hidden states of the top stacked LSTM layers as
\small
\begin{align*}
H^q = \mathrm{LSTM}(\phi(q)) = (h^q_1, \cdots, h^q_n)\\ 
H^t = \mathrm{LSTM}(\phi(t)) = (h^t_1, \cdots, h^t_m)
\end{align*}
\normalsize
Inspired by natural language understanding proposed in~\cite{wang2019cross, DBLP:conf/iclr/SeoKFH17}, we build an extra LSTM layer on $H^q$ with attention over $H^t$ as the follows
\small
\begin{align*} 
d_{i} &= \mathrm{LSTM} \left( [h^t_i,\beta_{i-1}], d_{i-1} \right) \\
e_{ij} &= v^T \Tanh(W_0 H^t + W_1 h_j^q + W_2 d_i) \\
\alpha_{ij} &= {e_{ij}} \large/ {\sum_{j'} e_{ij'}} \\
\beta_i &= \sum^n_{j=1} \alpha_{ij}  h_j^q
\end{align*}
\normalsize
$d_0 = \mathbf{0}$. $W_0$, $W_1$, $W_2$, $v$, and $U$ are model parameters.
Here $i$ is the time step while enumerating $t$, and $j$ enumerates each token in $q$.
The final output $d_m$ is fed to a multi-layer perceptron (MLP) and then resized to a binary prediction.
If $t$ involves a sequence of tokens, we use bi-directional attentive flow as in~\cite{DBLP:conf/iclr/SeoKFH17} and compute bi-directional output $d_i =[\overrightarrow{d_i}, \overleftarrow{d_i}]$. 

With the attentive flow on type $t$ while reading the question, our spatial comprehension model is able to make the prediction based on the memory of the context. 
However, with the observation that, given a question ``{\it How many rivers in Mississippi}?'' and a POI type ``{\it river}'', the machine comprehension model is highly likely to produce a positive prediction (false prediction), since ``{\it river}'' is mentioned in the question by ``{\it rivers}'', but the model would fail to capture our intention to categorize the type of ``{\it Mississippi}''. In order to feed our intention into the model,
we insert special symbols (e.g., $\langle$@$\rangle$) to enclose the POI mentioned in the question.
 
\begin{figure}[h!]
    \vspace{-5pt}
    \resizebox{.5\textwidth}{!}{
    \begin{tabular}{c|c|c}
     Question & POI Type & Label\\
    \hline
    {\it How many states does the $\langle@\rangle$ Mississippi $\langle@\rangle$ run through?}   & State & False\\
    {\it How many states does the $\langle@\rangle$ Mississippi $\langle@\rangle$ run through?}  & City & False\\
    {\it How many states does the $\langle@\rangle$ Mississippi $\langle@\rangle$ run through?}  & River &True\\
    \end{tabular}
    }
    \vspace{-10pt}
    \caption{Spatial Comprehension Model Training Samples}
    \label{fig:spatial-comprehension-input}
    \vspace{-10pt}
\end{figure}

The corresponding model structure is shown in Figure~\ref{fig:spatial-comprehension-model}.
For the question shown in Figure~\ref{fig:spatial-comprehension-model}, to address the ambiguity of ``{\it Mississippi}'', we feed three records shown in Figure~\ref{fig:spatial-comprehension-input}.
Our spatial comprehension function is defined as follows:
\small
$$\textsc{SpatialComprehension}(q,p,t)$$
\normalsize
where $q$ is the question, $p$ is the ambiguous phrase, and $t$ is the meaning of the phrase. 
If \textsc{SpatialComprehension} returns true, the semantic meaning of $p$ is identified as $t$.


\subsection{Spatial Semantics Injection}

\begin{figure}[b]
    \resizebox{.5\textwidth}{!}{
    \begin{tabular}{|c|l|}
    \toprule
     Question $q$&  {\it  What is the population of San Antonio} ?\\
    \hline
      Keyword & \multirow{2}{*}{\it What is the \textcolor{blue}{population} of \textcolor{blue}{San Antonio} ?}\\
      Detection & \\
    \hline
      Symbol $q'$&  {\it what is the \textcolor{blue}{$\langle k0\rangle$ population $\langle eok\rangle$} of}\\
      Injection & {\it \textcolor{green}{$\langle k1\rangle$ cityid $\langle eok\rangle$} \textcolor{blue}{$\langle v0\rangle$ San Antonio $\langle eov\rangle$}} ?\\
    \hline \hline
      \multicolumn{2}{|c|}{Seq2seq Model}\\
    \hline \hline
      Output $l'$ & \texttt{answer(A,\textcolor{blue}{$\langle k0 \rangle$}(B,A),const(B,\textcolor{green}{$\langle k1 \rangle$}(\textcolor{blue}{$\langle v0 \rangle$}))) }\\
    \hline
      Recover $l$ & \texttt{answer(A,\textcolor{blue}{population}(B,A),const(B,\textcolor{green}{cityid}(\textcolor{blue}{San Antonio})))} \\  
    \bottomrule
    \end{tabular}
    }\\
    \vspace{.1cm}
    \resizebox{.5\textwidth}{!}{
    \hspace{-10pt}
    \includegraphics{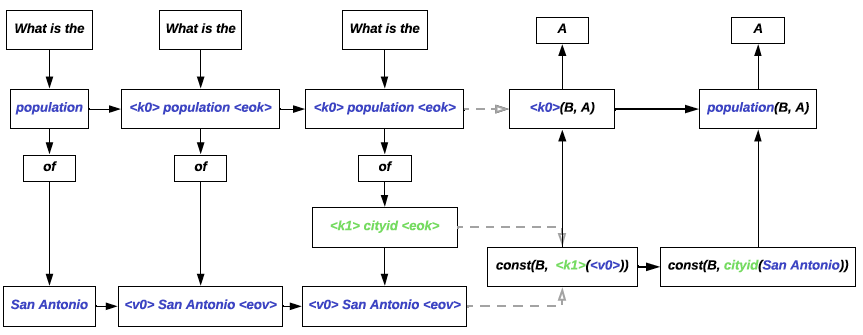}
    }
    \vspace{-15pt}
    \caption{An example of Information Injection Format with Type Feeding}
    \label{fig:injection2}
\end{figure}

Now we are able to understand the context and recognize the meaning of each ambiguous spatial phrase correctly through the spatial comprehension model. The question is how to inject the external information to the main seq2seq model.
We propose an injection strategy shown in Figure~\ref{fig:injection2}. The general idea is to insert symbols into the question at the locations before and after every spatial phrase to emphasize its semantics, then feed the inserted question into a seq2seq model. 

The first step is to search for components in the question that need a spatial semantics injection--i.e., ambiguous spatial semantics.
For example, in the question ``{\it  What is the population of San Antonio}?'', we do not feed extra information for every word, instead, we only focus on spatial information or tokens that could be shown in the corresponding logic form (e.g., keywords).
In other words, we only care about the tokens in the NL question that contribute to its logic form. The tokens such as question word ``what'' and stop words ``is'' ``the'' ``of'' do not contain the question's information, thus are not annotated. 

For the ambiguity of spatial phrases, we believe it is necessary to feed the meaning of the phrase in the question. For example, we feed the \textbf{Type of POI} in the question to address the POI ambiguity.
We present our \textit{Information Injection Format With Type Feeding} in Figure~\ref{fig:injection2}. We will validate in the experiment section that our type feeding improves the accuracy dramatically.

We propose a general purpose automatic injection algorithm shown in Algorithm \ref{alg:spatial_injection}. For the input natural language question $q$, after we recognize the meaning (e.g., type) $t_i$ of each phrase $p_i$ (e.g., POI) correctly through spatial comprehension model, for each pair of <$p_i$, $t_i$> $\in$ $<P,T>$, we will insert the type $t_i$ before the $p_i$ in the input question $q$. Also, for $p_i$ $\in$ $P$, we store the symbol $sym$ of each phrase $p_i$ in $s2p$, which will be used later in the query recovery. For example, as Figure \ref{fig:injection2} shows, the symbol for the phrase ``{\it san antonio}'' is $\langle v0\rangle$. Then the <$\langle v0\rangle$, $san$ $antonio$> is stored in $s2p$ which will later be used for the recovery of $\langle v0\rangle$ in the output logic form query.

\begin{algorithm}[h!]
\caption{Spatial Semantics Injection} 
\begin{algorithmic}[1]
\Function{\textsc{SpatialInjection}}{$q$, $V$, $P$}
      \State $s2p$ = $\emptyset$; \Comment{Symbol phrase mapping.}
      \State $index_v$ = $0$; \Comment{Value Symbol index.}
      \State $index_k$ = $0$; \Comment{Keyword symbol index.}
      \State $q'$ = $q$;
      \ForAll{$\langle p_q, c \rangle \in P$} \Comment{Iterate each matched pairs.}
        \If{$c$ is a keyword}
            \State $sym$ = `$k$'+$index_k$;
        \ElsIf{$c$ is a value}
            \State $sym$ = `$v$'+$index_v$;
            \State Search for $c$.$types$ from $V$;
            \If {|$c$.$types$|>1}\Comment{$c$ is an ambiguous spatial phrase}
                \State $T$ = $p.types$\;
                \ForAll{$t \in T$}
                    \If {\textsc{SpatialComprehension}$(q, p, t)$ is $True$}
                        \State $c$.$type$ = $t$;
                    \EndIf
                \EndFor
            \ElsIf {|$c$.$types$|==1} \Comment{$c$ is not an ambiguous spatial phrase}
                \State $c$.$type$ = $c$.$types[0]$;
            \EndIf
            \State Insert $c$.$type$ to $q'$ (using symbol `$k$'+$index_k$)\;
        \EndIf
        \State $index_k$ = $index_k$ + 1; 
        \State $index_v$ = $index_v$ + 1; 
        \State $s2p$.add($\langle$$sym$, $c\rangle$);
        \State Insert $sym$ to $q'$;
        
      \EndFor
     \State \textbf{Return} $q'$, $s2p$;
\EndFunction
\end{algorithmic}
\label{alg:spatial_injection}
\end{algorithm}

Figure~\ref{fig:injection2} presents our detailed injection format. For a phrase or token in the question that is identified as a keyword (e.g., population), that phrase or token will be enclosed with $\langle ki\rangle$ and $\langle eok\rangle$. For values such as ``{\it San Antonio}'' that appear in the question, we enclose them with $\langle vi\rangle$ and $\langle eov\rangle$ where 
$\langle eok\rangle$ represents ``\textsf{end of keyword}'' and
$\langle eov\rangle$ represents ``\textsf{end of value}''.
Note that we use the spatial databases and the grammar of executable queries to identify keywords and values without referring to the ground truth. Here $i$ indicates it is the $i$-th spatial semantics that is injected. For a value $\langle v\rangle$, if ambiguity exists, we predict its spatial meaning using spatial comprehension model and feed the spatial semantics into the question using the symbol $\langle k\rangle$.

As shown in Figure~\ref{fig:injection2}, since the output of the seq2seq model involves symbols that are inserted into the question which need to be transformed to its original literal form, we propose a query recovery model. The detailed algorithm will be presented in Section~\ref{sec:query_recover}.

Following the aforementioned example in spatial comprehension $q$ = ``{\it How many states does the Mississippi run though}?'', if we do not address the ambiguity problem and rely on the seq2seq model to infer the spatial meaning of ``{\it Mississippi}'',  $q'$ will be ``{\it How many $\langle k0 \rangle$ states $\langle eok \rangle$ does the $\langle v0 \rangle$ Mississippi $\langle eov \rangle$ run through}?''. After translated by the seq2seq model, the recovered query is likely to be \texttt{\small answer(A, count(B, (state(B), const(C,\textcolor{red}{stateid}(mississippi)), traverse(C,B)), A))} \normalsize  since ``{\it states}'' is mentioned in the question. Our model is fundamentally built upon a seq2seq translation model, where the context is transformed to a weighted sum of all the tokens, in which ``{\it states}'' will be embedded as part of the context, and the model is easy to be confused and outputs ``{\it Mississippi}'' as a state name. We will mention in Section~\ref{sec::model} later that, since the output vocabulary size is much smaller than the input vocabulary size, and most of the tokens in output appear in the input as well, we adopt Copying Mechanism~\cite{DBLP:conf/acl/JiaL16}, where the output token has a higher chance to be copied from the input sequence. Copying Mechanism makes the ambiguity harder to address, such as in the aforementioned question, ``{\it river}'' does not appear in the input question, the model has a higher probability to copy ``{\it state}'' from the input sequence. Thus we need to insert $`riverid'$ before the word ``{\it Mississippi}'' in the input sentence to help the model make the translation. The final input will be ``{\it How many $\langle k0 \rangle$ states $\langle eok \rangle$ does the $\langle k1 \rangle$ riverid $\langle eok \rangle$ $\langle v0 \rangle$ Mississippi $\langle eov \rangle$ run through}?''.

\section{Translate \& Recover}

\subsection{Translation Model}\label{sec::model}
Since seq2seq models have been widely adopted in translation tasks, and our NLI task is simpler than a translation task due to small vocabulary size.
We believe a seq2seq model is able to capture the logic and the spatial structure of the question as long as it is able to understand the entities that are mentioned in the question. So we adopt a seq2seq model with copying mechanism following~\cite{wang2018transfer}.
$$l' = \textsc{seq2seq}(q')$$

\subsection{Query Recovery Model}\label{sec:query_recover}
\begin{algorithm}[b]
\caption{Symbol Recovery} 
\begin{algorithmic}[1]
\Function{\textsc{Recover}}{$l'$, $s2p$}
     \State $l$ = $l'$;
     \ForAll {$\langle sym, c\rangle$ $\in$ $s2p$} 
        \State $l$.replace($sym$,$c$);
     \EndFor
     \State \textbf{Return} $l$;
\EndFunction
\end{algorithmic}
\label{alg:symbol_recover}
\end{algorithm}

We detail our strategy of query recovery through Algorithm~\ref{alg:symbol_recover}, whose inputs are $l'$, the output of translation model, and $s2p$, the symbol-phrase pairs detected by spatial semantics injection model (Algorithm~\ref{alg:spatial_injection}). Just as Figure~\ref{fig:injection2} shows, the output $l'$ of the seq2seq translation model is a sequence of the symbol, for example, $\langle k0 \rangle$, $\langle k1 \rangle$ and $\langle v0 \rangle$ here. Then we need to recover the output logic form query. For each pair $\langle sym, c\rangle$ $\in$ $s2p$, every symbol $sym$ $l'$ needs to be replaced by the original phrase $c$. After replacing all symbols in $l'$, we finally get the output logic form. In Figure \ref{fig:injection2}, for output $l'$, we replace the symbols $\langle k0 \rangle$, $\langle k1 \rangle$ and $\langle v0\rangle$ by their corresponding phrase ``{\it population}'',``{\it cityid}'' and ``{\it San Antonio}'' based on the pairs in $s2p$. After recovery, we finally get the right output logic form.

\section{Data Augmentation By Shuffling}\label{sec::aug}
Just as mentioned before, one of the Challenges right now is the lack of training set. The sparsity of training data causes two problems: 1). the semantic structures of questions are sparse;
2). the data entities mentioned in the questions are inadequate. Problem 2 can be simply addressed by replacing data entities. However, addressing problem 1 is non-trivial. So we propose to shuffle the prepositional phrases to augment the semantic structures of the training set.

We propose our unique augmentation strategy as follows: 
If a question has a prepositional phrase ($PP$ as a POS tag), and the question can be decomposed as $q = q^{\mathrm{prefix}} | q^{\mathrm{PP}}$ or $q = q^{\mathrm{PP}}|q^{\mathrm{suffix}}$, where $q^{\mathrm{prefix}}$ are the words placed before the prepositional phrase and $q^{\mathrm{suffix}}$ are the words placed after the prepositional phrase, we will shuffle the position of the prepositional phrase. Note that we only consider the questions that start with a prepositional phrase or end with a prepositional phrase.

Consider the example in Figure~\ref{fig:augmentation}, for the query \textit{"Which states does the Mississippi river run through"}, the format of the question is $q=q^{\mathrm{prefix}}|q^{\mathrm{PP}}$, and $q^{\mathrm{prefix}}$=``{\it Which states does the Mississippi river run}'' and $q^{\mathrm{PP}}=$``{\it through}''. By exchanging $q^{\mathrm{prefix}}$ and $q^{\mathrm{PP}}$, we can get a new sentence ``{\it Through which states does the Mississippi river run}'' and the meaning of the new query remains the same. 
Also, for the other question ``{\it In what state is Mount Mckinley }'', $q=q^{\mathrm{PP}}|q^{\mathrm{suffix}}$, $q^{\mathrm{PP}}$=``{\it In what state}'', and $q^{\mathrm{suffix}}$=``{\it is Mount Mckinley}''.
we can shuffle the position of the prepositional phrase to get a new sentence.

\begin{figure}[h!]
    \vspace{-5pt}
    \centering
    \resizebox{.4\textwidth}{!}{
    \begin{tabular}{|cl|}
    \toprule
     \multirow{2}{*}{$q$} &  {\it \underline{Which states does the Mississippi river run} \underline{through}} ?\\
     & \hspace{2.3cm} $q^{\mathrm{prefix}}$ \hspace{2cm} $q^{\mathrm{PP}}$\\
    \hline
     \multirow{2}{*}{Augment}   &  {\it \underline{Through} \underline{which states does the Mississippi river run}} ? \\
      &  \hspace{0.2cm} $q^{\mathrm{PP}}$ \hspace{2.3cm} $q^{\mathrm{prefix}}$ \hspace{2cm}\\
    \hline
     \multirow{2}{*}{$q$}  & {\it \underline{In what state} \underline{is Mount Mckinley}} ?  \\
     &  \hspace{0.5cm} $q^{\mathrm{PP}}$ \hspace{1.3cm} $q^{\mathrm{suffix}}$ \\
    \hline
     \multirow{2}{*}{Augment} & {\it \underline{Mount Mckinley is} \underline{in what state}} ?  \\
     &  \hspace{.5cm} $q^{\mathrm{suffix}}$  \hspace{1.3cm} $q^{\mathrm{PP}}$ \\
    \bottomrule
    \end{tabular}
    }
    \vspace{-10pt}
    \caption{Two examples of data augmentation}
    \label{fig:augmentation}
    \vspace{-10pt}
\end{figure}

\section{Experimental Validation}

\subsection{Experimental Settings}

\noindent\textbf{Configuration}
All our experiments are conducted on a machine equipped with 2 Intel CPU E5-2670 v3 running at 2.3GHz with 256GB of RAM and 2 NVIDIA Tesla K80 GPUs. 

\noindent\textbf{Dataset}
To evaluate the effectiveness of our system, we performed an experimental evaluation on dataset Geoquery and Restaurants.~\footnote{Our code is publicly available at https://github.com/VV123/SpatialNLI}  

\begin{itemize}
\item \textit{Geoquery}~\cite{zelle1996learning} is a collection of 880 natural language questions and corresponding executable database query pairs about U.S. geography. The answers in this dataset are defined in $\lambda$-calculus logical form. We follow the standard training-test split to that of \cite{DBLP:conf/uai/ZettlemoyerC05}, of which the dataset was divided into 600 training examples and 280 test examples respectively.  As \cite{DBLP:conf/acl/JiaL16}, \cite{DBLP:journals/coling/LiangJK13} and \cite{wang2015building}, we determine its $Acc$ based on the denotation match.
\item \textit{Restaurant} (Rest)~\cite{tang2000automated, DBLP:conf/iui/PopescuEK03} is a dataset with 251 question-answer pairs about restaurants, their food types, and locations. The questions are all human natural language and the answers are in $\lambda$-calculus logical form.
\end{itemize}

\subsection{Data Augmentation}

We not only propose our new data augmentation strategy by shuffling in Section~\ref{sec::aug}, but also adopt a data augmentation strategy that is based on the recombination~\cite{DBLP:conf/acl/JiaL16} of a sentence itself. 

For example, as in Figure \ref{fig:augmentation2}, for the query ``{\it What is the highest point in Florida?}'', we can simply identify that the word ``{\it Florida}'' is the name of a state based on the spatial database. Given this example, we change it to new questions, in which the word ``{\it Florida}'' is replaced by the name of other states in the database. Here, the word ``{\it Florida}'' is replaced by ``{\it Rhode Island}''. For the second example in Figure \ref{fig:augmentation2}, for the query ``{\it What is the highest point in Florida?}'' and the query ``{\it What state has the smallest population density?}'', we can infer that the entire expression of the second sentence could map to the word ``{\it Florida}'' in the first query since this query is asking about one state. Then we can generate one new question by replacing the word ``{\it Florida}'' with the second sentence. For the third example, the two queries, ``{\it What state has the largest population?}'' and ``{\it What state has no rivers?}'', are both asking about one state, so we combine them together to generate a new query.

\begin{figure}[t]
    \centering
    \resizebox{.5\textwidth}{!}{
    \begin{tabular}{|ccl|}
    \toprule
    \multirow{2}{*}{Type 1}& Original &  {\it What is the highest point in Florida}?\\
    & Augment   &  {\it What is the highest point in Rhode Island}? \\
    \hline
    \multirow{4}{*}{Type 2}&\multirow{2}{*}{Original} & {\it What is the highest point in Florida}? \\
    & & {\it what state has the smallest population density}?\\
    &\multirow{1}{*}{Augment} & {\it What is the highest point in state that has the smallest population density}?\\ 
    \hline
     \multirow{4}{*}{Type 3}& \multirow{2}{*}{Original} & {\it what state has the largest population}? \\
    & & {\it what state has no rivers}?\\
    & \multirow{1}{*}{Augment}  &  {\it what state has the largest population and has no rivers}?\\
    \bottomrule
    \end{tabular}
    }
    \vspace{-10pt}
    \caption{Three examples of data augmentation}
    \label{fig:augmentation2}
\end{figure}

\subsection{Spatial Comprehension Model}

\begin{table}[b]
	\centering
	\resizebox{0.3\textwidth}{!}{
		\begin{tabular}{|p{1.5cm}|p{1cm}p{1cm}p{1cm}|}
			\toprule
			 Dataset & & Train & Test\\
			\hline
			 \multirow{2}{*}{Geoquery}&$Acc_{rcd}$ & 97.4\%& 91.9\%  \\
			        &$Acc_{qu}$ & 98.3\%& 98.1\% \\
			 \hline
		     \multirow{2}{*}{Rest(aurant)} &$Acc_{rcd}$ & 100.0\%& 100.0\%  \\
			        &$Acc_{qu}$ & 100.0\%& 100.0\% \\
			\bottomrule
		\end{tabular}
	}
	\caption{Spatial Comprehension Model evaluation.}
	\label{table:spatial-comprehension}
\end{table}

We preprocess the dataset for spatial comprehension model so that each record contains (1) A question with each POI phrase enclosed with symbols (e.g., $\langle@\rangle$) indicating the attentive position;
(2) A POI type (e.g., River, State, and City.).

For a question ``{\it How many states does the Mississippi run through?}'' with one ambiguous POI ``Mississippi'', we have the three records as shown in Figure~\ref{fig:spatial-comprehension-input}.
To balance the positive and negative samples in the training set, we replicate positive samples. For samples in Figure~\ref{fig:spatial-comprehension-input}, we replicate positive samples by $2$ times.

We run experiments with 200 hidden units and 300-dimensional pre-trained Glove embedding. We minimize the cross entropy using Adam Optimization Algorithm.
We evaluated the performance of our spatial comprehension model in Table~\ref{table:spatial-comprehension}. $Acc_{rcd}$ represents the percentage of correctly predicted records. $Acc_{qu}$ represents the percentage of correctly predicted questions where all POIs are recognized correctly.
For the example in Figure~\ref{fig:spatial-comprehension-input}, the total number of samples for $Acc_{rcd}$ is 3, and the total number of samples for $Acc_{qu}$ is 1. Even the training objective function is to optimize $Acc_{rcd}$. In fact $Acc_{qu}$ is what we are trying to optimize, and we prove that $Acc_{rcd}$ and $Acc_{qu}$ are optimized simultaneously.

We evaluate on Geoquery and Rest datasets, respectively (shown in Table~\ref{table:spatial-comprehension}).
Test $Acc_{qu}$ is $98.1\%$ for Geoquery, and $100.0\%$ for Restaurant data, respectively. All the $Acc_{qu}$ is not less than $Acc_{rcd}$. In other words, our spatial comprehension model is able to recognize the spatial semantics with high confidence. 

\subsection{Evaluation}

For the encoder and the decoder of our seq2seq model, we use one layer of Gated Recurrent Unit (GRU) with hidden size of $800$ and $800*2$, respectively.
The input and output of both encoder and decoder share the same embedding layer, which is initialized with 300-dimensional pre-trained Glove embedding.
Special symbols inserted (e.g., $k_1$ and $v_1$) are treated as special tokens; they are represented by the concatenation of an embedding of the symbol type (e.g, $k$ and $v$) and an index, where 
the embedding of the symbol type and the index are randomly initialized with $150$-dimension (the concatenation has a dimension of $300$).
The other unknown token is initialized with a $300$-dimension random vector. 
For training, we use gradient clipping with a threshold $5.0$, and for inference, we use beam search with width $5$.

\begin{table}[b]

	\centering
	\resizebox{0.4\textwidth}{!}{%
		\begin{tabular}{|c|lc|}
			\toprule
			Dataset & \multicolumn{2}{c|}{Geoquery}\\
			\midrule
			\multirow{7}{*}{Conventional}
			& ZC05 ~\cite{DBLP:conf/uai/ZettlemoyerC05}&
			79.3\% \\
			& ZC07~\cite{zettlemoyer2007online} & 86.1\%\\
			& UBL ~\cite{DBLP:conf/emnlp/KwiatkowksiZGS10}& 87.9\% \\
			& DCS+L~\cite{DBLP:journals/coling/LiangJK13}& 87.9\%  \\
			& FUBL ~\cite{DBLP:conf/emnlp/KwiatkowskiZGS11}& 88.6\% \\
			& ZH15 ~\cite{DBLP:conf/naacl/ZhaoH15} & 88.9\%  \\
			& KCAZ13~\cite{kwiatkowski2013scaling}& 89.0\% \\
			\midrule
			\multirow{4}{*}{Deep Model} & ASN~\cite{DBLP:conf/acl/RabinovichSK17} & 87.1\%  \\
            & SQL2TREE~\cite{DBLP:conf/acl/DongL16} & 87.1\% \\
            & TRANX~\cite{DBLP:conf/emnlp/YinN18} & 88.2\% \\
            & JL16~\cite{DBLP:conf/acl/JiaL16} & 89.3\%\\
			\midrule
			\multirow{6}{*}{Ours}& SpatialNLI   & $\mathbf{90.4\%}$ \\
			& \quad\quad -- Copy Mechanism & 88.9 \% \\
		    & \quad\quad -- Spatial Comprehension & 86.4 \%  \\
		    & \quad\quad -- Type Feeding & 85.0 \%  \\
			& \quad\quad -- Data Augmentation & 83.2 \% \\
			& \quad\quad -- Information Injection & 82.9 \%  \\
			\bottomrule
		\end{tabular}
	}\\
	\label{table:expr_comparison_geo}
    (a)\\
    \vspace{5pt}
	\centering
	\resizebox{0.4\textwidth}{!}{%
		\begin{tabular}{|c|lc|}
			\toprule
			Dataset & \multicolumn{2}{c|}{Restaurant}\\
			\midrule
			\multirow{2}{*}{Conventional}
			& PEK03~\cite{DBLP:conf/iui/PopescuEK03} & 97.0\% \\
			& TM00~\cite{tang2000automated}   & 99.6\% \\
			\midrule
			\multirow{1}{*}{Deep Model} &  FKZ18~\cite{DBLP:conf/acl/RadevKZZFRS18} & 100.0\%\\
			\midrule
			\multirow{6}{*}{Ours}&  SpatialNLI   & $\mathbf{100.0}$\% \\
			& \quad\quad -- Spatial Comprehension & 96.1 \% \\
			& \quad\quad -- Copy Mechanism & 94.1 \% \\
			& \quad\quad -- Data Augmentation & 92.2 \% \\
			& \quad\quad -- Type Feeding & 70.6 \% \\
			& \quad\quad -- Information Injection & 60.8 \% \\
			\bottomrule
		\end{tabular}
	}\\
	(b)\\
	\caption{``--'' means the removal of each component. The accuracy is measured as denotation match~\cite{DBLP:conf/acl/JiaL16} on test set.}
	\label{table:expr_comparison_rest}
\end{table}

\begin{figure*}[h!]
    \centering
    \resizebox{.9\textwidth}{!}{
    \begin{tabular}{|c|c|l|}
    \toprule
      & $q$ &  {\it How many states does the \textcolor{red}{Mississippi} run through} ?\\
     \hline
     \hline
     \multirow{3}{*}{+SC} & \multirow{1}{*}{$q'$} &{\it $\langle k0 \rangle$ How many $\langle eok \rangle$ $\langle k1 \rangle$ states $\langle eok \rangle$ does the \textcolor{blue}{$\langle k2 \rangle$ riverid $\langle eok \rangle$} \textcolor{red}{$\langle v0 \rangle$ Mississippi $\langle eov \rangle$} run $\langle k3 \rangle$ through $\langle eok \rangle$} ?\\
     \cline{3-3}
     &\multirow{1}{*}{Infer} & \multirow{1}{*}{\texttt{answer(A,$\langle k0 \rangle$(B,($\langle k1 \rangle$(B),const(C,\textcolor{blue}{$\langle k2 \rangle$}(\textcolor{red}{$\langle v0 \rangle$})),$\langle k3 \rangle$(C,B)),A))}}\\
     \cline{3-3}
     &\multirow{1}{*}{Recover} & \texttt{answer(A,count(B,(state(B),const(C,\textcolor{blue}{riverid}(\textcolor{red}{Mississippi})),traverse(C,B)),A))}\\ 
    \hline     
    \multirow{3}{*}{-SC} &\multirow{1}{*}{$q'$} &{\it $\langle k0 \rangle$ How many $\langle eok \rangle$ $\langle k1 \rangle$ states $\langle eok \rangle$ does the \textcolor{blue}{$\langle k2 \rangle$ stateid $\langle eok \rangle$} \textcolor{red}{$\langle v0 \rangle$ mississippi $\langle eov \rangle$} run $\langle k3 \rangle$ through $\langle eok \rangle$} ?\\
     \cline{3-3}
     &\multirow{1}{*}{Infer} & \multirow{1}{*}{\texttt{answer(A,$\langle k0 \rangle$(B,($\langle k1 \rangle$(B),const(C,\textcolor{blue}{$\langle k2 \rangle$}(\textcolor{red}{$\langle v0 \rangle$})),$\langle k3 \rangle$(C,B)),A))}}\\
     \cline{3-3}
     &\multirow{1}{*}{Recover} & \texttt{answer(A,count(B,(state(B),const (C,\textcolor{blue}{stateid}(\textcolor{red}{Mississippi})),traverse(C,B)),A))}\\
    \bottomrule
    \end{tabular}
    }
    \vspace{-10pt}
    \caption{Spatial Comprehension Case Study. *+SC means using Spatial Comprehension, -SC means without.}
    \label{fig:case3}
    \vspace{-10pt}
\end{figure*}

Table~\ref{table:expr_comparison_rest} presents our experiment results for (a) Geoquery dataset and (b) Rest(aurant) dataset. 
For Geoquery, compared with the previous models, our method outperforms the state-of-the-art. The conventional methods are overdependent on predefined templates and manually designed features, which have lower accuracy on the test set. For neural network-based methods such as ASN~\cite{DBLP:conf/acl/RabinovichSK17} and TRANX~\cite{DBLP:conf/emnlp/YinN18}, they convert word space to action space, which inevitably introduces transformation error. SQL2TREE~\cite{DBLP:conf/acl/DongL16} and JL16~\cite{DBLP:conf/acl/JiaL16} use seq2seq model as well,
but fail to address spatial semantics ambiguity. For Rest, the state-of-the-art achieves $100\%$ accuracy, which states the Rest dataset is an easier task than Geoquery. Our model exhibits excellent downward compatibility by achieving $100\%$ accuracy on Rest dataset.

To validate the performance of our system, several ablation experiments were conducted by the removal of
(1)  Copy Mechanism, (2) Spatial Comprehension Model, (3)  Data Augmentation, (4) Type Feeding and (5) Information Injection, respectively.
By the removal of the spatial comprehension model, we random guess the meaning (type) of ambiguous POI and inject it to the question. For removing information injection, we feed the original content to the model without inserting any symbols.
By the removal of type feeding, we conduct symbol injection but omit to inject the extra spatial information (e.g., $\langle k1\rangle$ cityid $\langle eok\rangle$ in Figure~\ref{fig:case1}).

First, we measure the contribution of the spatial comprehension mechanism to the overall performance of the model. We train and evaluate two models: one with the spatial comprehension model and one without. Training is done with data augmentation and information injection. In Table \ref{table:expr_comparison_rest}, for Geoquery and Restaurant, with the removal of spatial comprehension model, the denotation match accuracy drops $4\%$ for Geo and drops $3.9\%$ for Rest. Since only $19.3\%$ of the test set for Geoquery and $4\%$ of the test set for Restaurant has POI ambiguity problem, it is obvious that our machine comprehension model is able to resolve the majority of them.

As shown in Figure~\ref{fig:case3}, by comparing against the spatial database, ``{\it Mississippi}'' appears in two tables: River table and State table. Without spatial comprehension, if we are using random guess, ``{\it river}'' has only $50\%$ chance to be correctly categorized. The `+comprehension' in the figure means we use the spatial comprehension model and `-comprehension' is for the result without right understanding of ``{\it Mississippi}''. Without the spatial comprehension model, it is possible for the system to recognize the ``{\it Mississippi}'' as a state name. As the figure shows, once ``{\it Mississippi}'' is recognized as a state, it will insert ``\textit{stateid}'' in the input question and finally get a wrong result after recovery. One interesting thing is that the infer for `+comprehension' and  `-comprehension' are the same, both correct. This is because, for seq2seq model, it just outputs the result with $\langle ki \rangle$, not the specific word. Here the ``\textit{riverid}'' and ``\textit{stateid}'' are both replaced by $\langle k2 \rangle$. Thus we get the same infer result from seq2seq model. But after recovery, the result without spatial comprehension model is wrong.

Table \ref{table:expr_comparison_rest} shows that by removing type feeding, the accuracies drop $5.4\%$ on Geoquery and $29.4\%$ on Restaurant. The symbol injection significantly improves the accuracy of the Restaurant dataset since for most samples in the Restaurant dataset, one token in the input question always corresponds to multiple tokens/symbols in the output sequence, which relationships are hard for the seq2seq model to capture. 


As shown in Table \ref{table:expr_comparison_rest}, the information injection component improves test accuracy by $7.5\%$ on Geoquery and $39.2\%$ on Restaurant. When we stop injecting information into the natural language question, the seq2seq is not able to capture all the necessary information to infer correctly and suffers from a large accuracy decrease. A case study of our symbol injection strategy is shown in Figure~\ref{fig:case1} where a seq2seq model generates outputs token by token and a large number of entities involve a sequence of tokens.
Without symbol injection, the seq2seq model has to infer ``{\it San Antonio}'' token by token using two steps.
On the other hand, with symbol injection, the seq2seq model generates $v0$ as a representation of ``{\it San Antonio}'', which only requires one step. Our symbol injection format is able to replace a name entity composed of a sequence of tokens to a single symbol, which prevents wrong name entity caused by a long sequential generation.

\begin{figure}[t]
    \centering
    \resizebox{.5\textwidth}{!}{
    \begin{tabular}{|l|c|l|}
    \toprule
    \multirow{2}{*}{-SI} & $q$ &  {\it What is the population of San Antonio} ?\\
    \cline{2-3}
     & Infer & \texttt{answer(A, population(B,A), const(B,cityid(\textcolor{red}{San Jose}))) }\\
    \hline \hline
    \multirow{5}{*}{+SI}&  $q$ &  {\it What is the population of San Antonio} ?\\
    \cline{2-3}
     & \multirow{2}{*}{$q'$} & {\it what is the $\langle k0\rangle$ population $\langle eok\rangle$ of $\langle k1\rangle$ cityid $\langle eok\rangle$}\\               
     &  & {\it $\langle v0 \rangle$ San Antonio $\langle eov\rangle$} ?\\
     \cline{2-3}
     & Infer & \texttt{answer(A, $\langle k0\rangle$(B,A), const(B,$\langle k1\rangle$($\langle v0\rangle$)))}\\
     \cline{2-3}
     & Recover & \texttt{answer(A, population(B,A), const(B,cityid(\textcolor{blue}{San Antonio})))}  \\
    \bottomrule
    \end{tabular}
    }
    \vspace{-10pt}
    \caption{ A Case Study using Symbol Inject (Geoquery). * +/- SI means with/without Symbol Injection.}
    \label{fig:case1}
\end{figure}

We also jointly train both datasets in a shared model
compared with separate training, shown in Table~\ref{table:merge}. Jointly training achieves an accuracy of $90.7\%$. Our experiment results show that a shared model performs better than two separate models.

\begin{table}[h!]
	\centering
	\resizebox{0.3\textwidth}{!}{
		\begin{tabular}{|p{2cm}|p{1.5cm}p{1.5cm}|}
			\toprule
			Training & Geoquery  &Restaurant \\
			\hline
			Separately & 90.4\% & 100\%\\
			Jointly & $\mathbf{90.7}\%$ & $\mathbf{100\%}$\\
		    \bottomrule
		\end{tabular}
	}
	\caption{Evaluation of jointly training on denotation match.}
	\label{table:merge}
	 \vspace{-15pt}
\end{table}
\section{conclusion}
In this work, we propose an NLIDB applied for the spatial domain to convert natural language queries to structured queries executable by database. The main contribution of our work is to recognize the meaning of the ambiguous spatial phrases based on contextual interpretation and capture the semantic structure of the question by the seq2seq model with injecting spatial information. Our extensive experimental analysis
demonstrates the advantage of our approach over state-of-the-art methods.

\balance
\bibliographystyle{unsrt}

\bibliography{sample-bibliography}
\end{document}